\documentclass[final]{cvpr}

\usepackage{times}
\usepackage{epsfig}
\usepackage{graphicx}
\usepackage{amsmath}
\usepackage{amssymb}
\usepackage{bbm}

\usepackage[pagebackref=true,breaklinks=true,colorlinks,bookmarks=false]{hyperref}

\def\cvprPaperID{8} 
\def\confYear{AML-CV 2021}
\begin{document}

\title{Adversarial Evaluation of Multimodal Models\\ under Realistic Gray Box Assumptions} 

\author{
Ivan Evtimov\\
U. of Washington\thanks{Work done while at Facebook AI.}\\
\and
Russell Howes\\
Facebook AI\\
\and
Brian Dolhansky\\
Reddit$^*$ \\
\and
Hamed Firooz\\
Facebook AI\\
\and
Cristian Canton\\
Facebook AI\\
}

\maketitle

\begin{abstract}
This work examines the vulnerability of multimodal (image + text) models to adversarial threats similar to those discussed in previous literature on unimodal (image- or text-only) models.
We introduce realistic assumptions of partial model knowledge and access, and discuss how these assumptions differ from the standard ``black-box''/``white-box'' dichotomy common in current literature on adversarial attacks. Working under various levels of these ``gray-box'' assumptions, we develop new attack methodologies unique to multimodal classification and evaluate them on the Hateful Memes Challenge classification task. We find that attacking multiple modalities yields stronger attacks than unimodal attacks alone (inducing errors in up to 73\% of cases), and that the unimodal image attacks on multimodal classifiers we explored were stronger than character-based text augmentation attacks (inducing errors on average in 45\% and 30\% of cases, respectively). 
\end{abstract}

\section{Introduction}
Multimodal reasoning has become an important ingredient in classification tasks for integrity enforcement for content in ad networks and on social media~\cite{facebook2020keeping,youtube2020protecting,twitter2020update}.
Posts with text and images are often evaluated for illegal, harmful, or hateful content with models that process both language and visual information.
These use cases are subject to pressure from adversarial bad actors with ideological, financial, or other motivations to circumvent models that identify violating content.
For example, a politically extremist group attempting to post misleading or violent content without being detected and removed might attempt to circumvent such models.
A seller in an online marketplace may try to obscure a listing for counterfeit goods, drugs, or other banned products.

\begin{figure}[tb]
    \centering
    \includegraphics[width=\columnwidth]{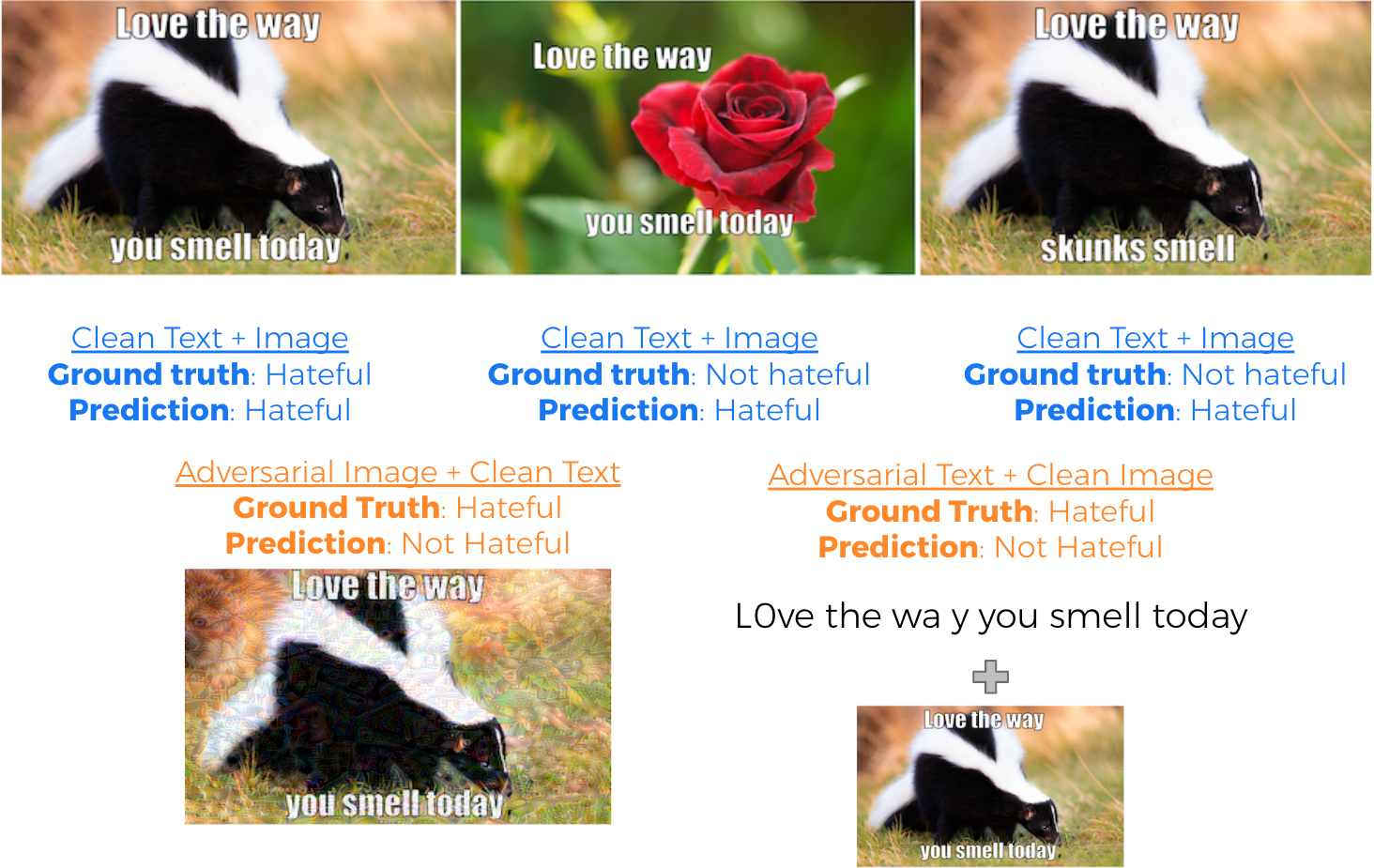}
    \caption{ 
        Examples of hateful and non-hateful memes in the Hateful Memes dataset~\cite{kiela2020hateful} and adversarial image and text inputs like the ones we generate. 
        Images originally available \href{https://www.drivendata.org/competitions/64/hateful-memes/}{here} and accessed on Oct 28th, 2020. Image above is a compilation of assets, including \copyright\:Getty Images.
    }
    \label{fig:hateful_memes_examples}
\end{figure}

In this work, we contribute to the understanding of how multimodal models may be vulnerable to threats observed against image-only and text-only models.
As a well-representative case study, we focus on the Hateful Memes Challenge and Dataset~\cite{kiela2020hateful}. 
This Challenge centers on the binary classification problem of labelling memes (images with embedded text) as ``hateful'' or ``non-hateful''. 
The notion of ``hatefulness'' in the Dataset is determined by how the concepts portrayed in the image interact with the meme text as shown in Figure~\ref{fig:hateful_memes_examples}.

Two unexplored points need to be addressed to better understand the risk to multimodal classification in practical settings. 
First, previous adversarial work typically assumes one of two threat models: ``white-box'' attacks with full access to a target model, and ``black-box'' attacks allowing only access to model inputs and outputs.
In practice, however, shared or inferred knowledge about such classifiers lies somewhere in between these two extremes.
Some classifiers may use public, off-the-shelf components (for example, object detector models) in their classification pipeline, while other information about the architecture is unknown.
This partial knowledge, while not as helpful to an attacker as full access to model weights, may still allow more powerful attacks than input/output access alone.

Second, it is not obvious how an image-based adversarial attack would impact a multimodal classifier compared to a unimodal, image-only classifier. Would the additional text input improve the robustness of the multimodal classifier, or would the multimodal classifier be \textit{less} robust if both the image and text components can be adversarially attacked?

This paper attempts to address this knowledge gap by quantifying how adversarial examples and text augmentations affect different variants of multimodal classification, under realistic threat models. 
Our major findings include:
\begin{itemize}
    \item Adversarial vulnerability does not really depend on the mechanism of combining text and image features (known as multimodal ``fusion''); models with different types of fusion fail under adversarial inputs at similar rates.
    \item Although the multimodal Hateful Memes models are heavily text-dependent, image-only attacks can impact classifier accuracy more than text-only attacks.
    \item Reusing public off-the-shelf components enables weaker adversaries to carry out successful attacks.
    In particular, models using pretrained object detectors without fine tuning are vulnerable to attack.
    \item Under ``gray-box'' assumptions, combined attacks against both modalities (image and text) are stronger than adversarially perturbing only one modality.
\end{itemize}

\section{Related Work}
\label{sec:related_work}
Machine learning has long been evaluated for security and robustness~\cite{dalvi2004adversarial,lowd2005adversarial,lowd2005good}. 
With the advent and wide deployment of neural networks for image and text processing in numerous practical scenarios, new threats have emerged as well.
Since 2013, the research community has focused extensively on \textit{adversarial examples}~\cite{szegedy2013intriguing}, images that a machine learning model misclassifies, even though they are visually similar (for humans) to benign images the model handles correctly.
Subsequent literature has been characterized by an ongoing attack/defense cycle~\cite{carlini2017towards,athalye2018obfuscated,carlini2017adversarial, carlini2017towards, tramer2020adaptive}. 
We apply the algorithm from  Madry \emph{et al.}~\cite{madry2017towards} in our work, and discuss its details in Section~\ref{sec:methodology}.
We also note that some work in this space has explored adversaries without access to the model~\cite{ilyas2018black,papernot2017practical} and we also draw on the lessons from the ``transferability'' research literature for our gray box attack models~\cite{liu2016delving}.

Natural language processing models have seen their share of attacks as well. 
Many approaches for generating adversarial text from a benign string use a two-step process that is repeated until an adversarial string is found.
In stage one, the algorithm selects the most ``important'' input token to modify; this can be done either with gradient information~\cite{cheng2020seq2sick,ebrahimi2017hotflip,li2018textbugger} or with masked queries~\cite{alzantot2018generating,garg2020bae,jin2019bert,li2020bert,pruthi2019combating,zang2020word}.
In the second step, the selected text is replaced with a suitable candidate; some works simulate typos in words~\cite{li2018textbugger}, others pick neighboring words in a semantically-aware embedding space~\cite{jin2019bert,ren2019generating}, and others use a language model (such as BERT~\cite{devlin2018bert}) to pick adversarial sentences that read ``naturally''~\cite{li2020bert}.
These attacks are standardized and implemented in a popular GitHub repository~\cite{morris2020textattack}.  

A limited set of works has explored the vulnerability of multimodal models but none (to our knowledge) has focused on the classification setting that we study.
There exist image-based attacks on VQA that produce answers of the attacker's choosing~\cite{sharma2018attend, xu2018fooling}.
Others have used adversarial modifications in an intermediary feature space in the training process to produce more generalizable models~\cite{gan2020large} but they have not produced images that generate those features.


\section{Methodology}
\label{sec:methodology}
Our goal for all examples in this section is to create memes that are either hateful but misclassified as non-hateful, or non-hateful but misclassified as hateful.
The memes generated by our attacks must also retain their original meaning, as judged by human users.
We use adversarial examples in the image domain and adversarial text augmentations for the text domain. 
We provide background on multimodal classification in Appendix~\ref{sec:background_multimodal} and further specify the different threat models that we evaluate in Appendix~\ref{sec:threat_model}.

\subsection{Attacks in the Image Domain}
In all cases, we adopt the projected gradient descent (PGD) algorithm from~\cite{madry2017towards} to modify the image while holding the text constant.
We provide full details of our attack implementations along with hyperparameter and model choices in Appendix~\ref{sec:more_methodology}.
Here, we summarize the 4 major types of attacks we consider.

\textbf{Full-Access and Dataset-Access Attacks}
In full-access scenarios, an adversary backpropagates through the exact model under attack while using a binary cross-entropy loss function with adversarial labels (opposite of the ground truth).
In the dataset-access case, the adversary cannot access the model they wish to fool.
However, they can train their own multimodal model and use it to generate adversarial examples in a full-access fashion.
As with previous research on adversarial examples transferability~\cite{liu2016delving}, we find that this method yields strong attacks.

\begin{figure}[tb]
    \centering
    \includegraphics[width=\columnwidth]{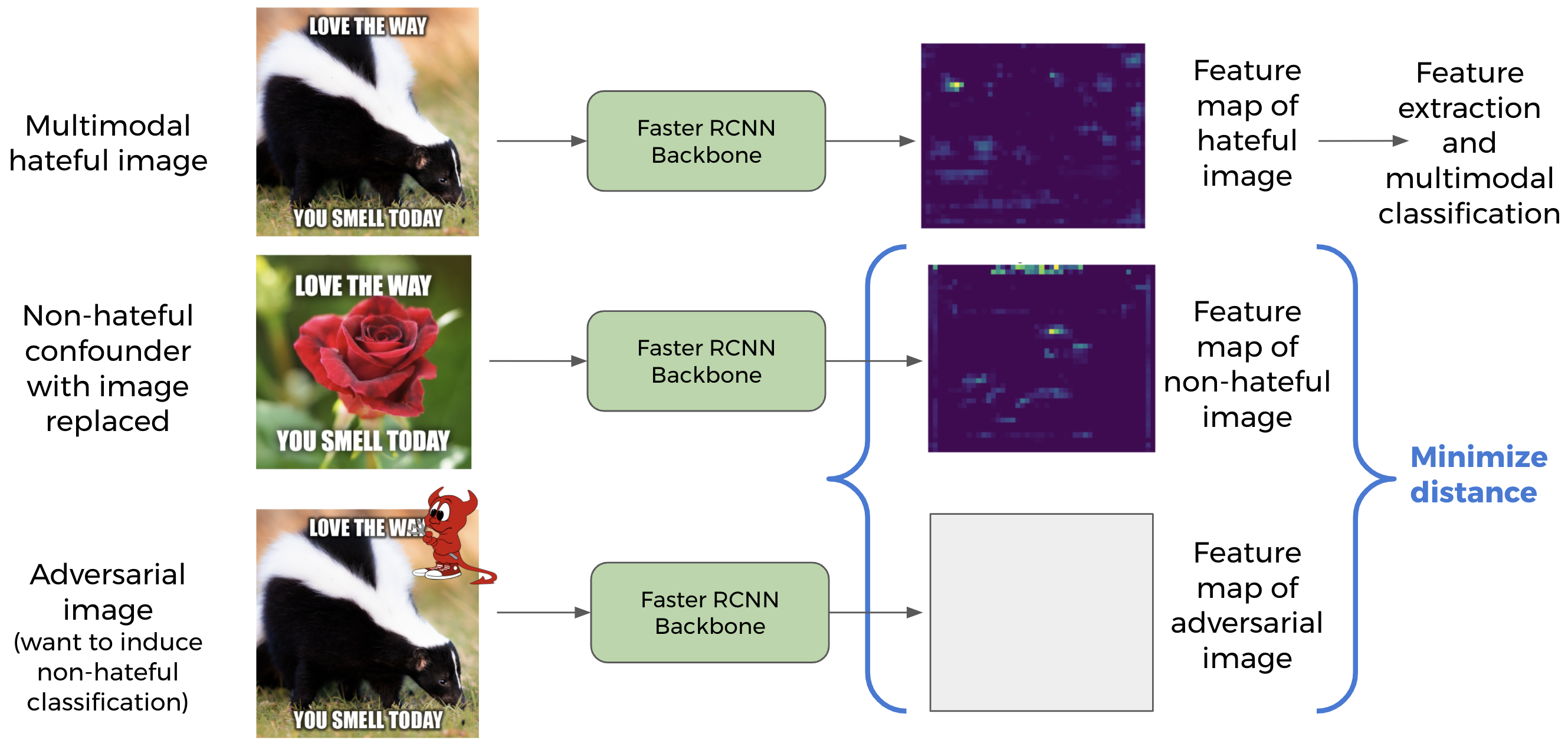}
    \caption{ 
        Illustration of one of the ``gray box'' image-based strategies we explore.
    }
    \label{fig:matching_text}
\end{figure}

\textbf{Feature Extractor-Access Attacks}
When multimodal classification is based on image region features, adversaries have access to a public, off-the-shelf component used in the prediction (such as the Faster R-CNN object detector~\cite{ren2015faster}).
In order to choose adversarial features without using gradient or query information from the multimodal model, adversaries can aim to disrupt interactions between the two modalities.
Thus, an adversary seeking to make a hateful meme be classified as non-hateful could add perturbations that shift the features that the image of the hateful meme produces to the features of the image from the non-hateful confounder.
We visualize this idea in Figure~\ref{fig:matching_text}.

\textbf{No-Access Attacks by Technically Savvy Adversaries}
In some scenarios, the adversary does not have access to the weights of any model trained on this task and cannot query them.
They can instead carry out a so-called ensemble attack on a set of standard public computer vision classifiers. 
In this kind of attack, adversarial examples are generated with a white-box attack that averages the gradient from $n$ models. 

\textbf{No-Access Attacks by Adversaries with No Expertise}
In scenarios with no query access, the adversary cannot use any gradient information (approximated or not) to generate their attacks. 
They can, however, introduce arbitrary modifications to the image that preserve its message. 
While there are multiple ways to do this, we use Gaussian noise of the same magnitude as our adversarial noise as a stand-in.

\subsection{Attacks in the Text Domain}
\label{sec:adversarial_text}
We also perform text-based attacks while maintaining the original image to study the importance of this modality to the adversarial robustness of multimodal classification.
We introduce two kinds of text attacks: guided (corresponding to full-access and dataset-access image attacks) and random augmentations (corresponding to no-access scenarios).
In both cases, we use the following set of augmentations: inserting emojis, replacing characters with their ``fun fonts'' equivalent, replacing letters with random other letters or with random unicode characters, inserting typos, splitting words.
Full details of our algorithm are given in Appendix~\ref{app:adversarial_text}.

\section{Experimental Results and Observations}
\label{sec:experiments}
We apply each of the methods described in Section~\ref{sec:methodology} to generate adversarial examples and adversarially augmented text.
In each case, we start with an example in the Hateful Memes test set and modify it with the goal that the prediction for that example changes to the opposite class from its ground truth.
However, recall from Figure~\ref{fig:mm_models} that the multimodal models we work with only achieve 60-70\% accuracy and 0.6-0.7 ROC AUC on the test set. 
In other words, the models misclassify a significant portion of the test set, even without adversarial modifications.

To measure only the effect of our attacks, we report the \textit{proportion of memes that were correctly classified when clean but misclassified when adversarial} out of the memes that were classified correctly to begin with. 
Formally, for a model $f$ and a dataset of memes $\mathcal{D} = \{(x_i, y_i)\}$ where $x_i$ is a ``clean'' meme containing text and image and $y_i$ is its ground truth label, we generate adversarial memes $x_i^{\text{adv}}$. 
Then, we report:
\[
     \frac{
        \sum_{i} \mathbbm{1}{ \{f(x_i) = y_i \text{ and }  f(x_i^{\text{adv}}) \neq y_i \}}
    }{
          \sum_{i} \mathbbm{1}{\{ f(x_i) = y_i \} }
    }
\]

\subsection{Image Adversarial Examples}
\label{sec:exp_image}

\begin{figure}[tb]
    \centering
    \includegraphics[width=\columnwidth]{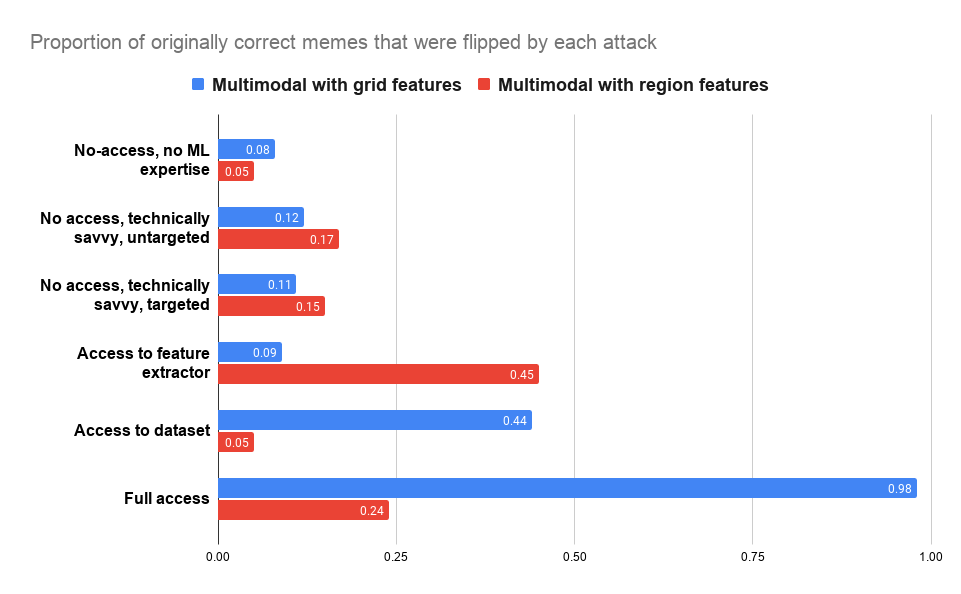}
    \caption{ 
        Averaged proportion of memes that were originally correct but misclassified after adversarial modifications of the image.
        For each of the two categories of multimodal models, we average the metric for each individual model over the three different models in that category.
    }
    \label{fig:image_results}
\end{figure}

For image attacks, we report results by averaging over all models in the two categories of image feature extractors: grid features and region features.
Results are given in Figure~\ref{fig:image_results}.

To put our attacks in context, first observe the least powerful adversary (inserting Gaussian noise) can flip only around 5 to 8\% of memes while the full-access adversary (who backpropagates through the model) can flip 98\% of memes that were originally correct.\footnote{
  With region feature models, we did not achieve 100\%; see App.~\ref{sec:full_access}.
}
Next, ``gray box'' attacks with similar effectiveness exist for both grid feature models and region feature models.
Dataset-access adversaries can flip 44\% of originally correct memes on average and adversaries with access to the region features extractor can achieve 45\%.
Additionally, technically savvy adversaries with no access can also induce errors in both categories of models at about the same rates (12-17\%).
This suggests that the nature of the image feature extraction might not make a model fundamentally more robust. 

However, there is an important asymmetry in which attacks perform well on which models.
Because grid feature models include retraining the image feature extractor, it takes a strictly more powerful adversary (one with access to the training dataset) to achieve the same goal.
This suggests that region feature extractors are, in practice, more at risk of adversarial compromise because an exact component of their operational pipeline is freely available to adversaries. 

\subsection{Text-Only Attacks}
\label{sec:exp_text}
\begin{figure}[tb]
    \centering
    \includegraphics[width=\columnwidth]{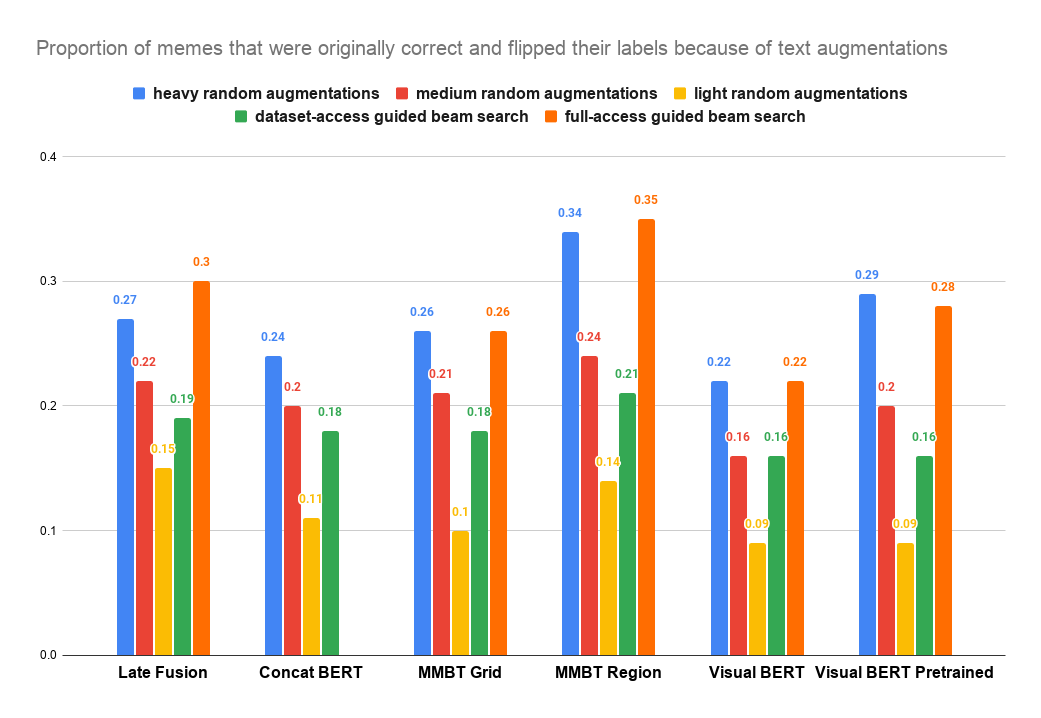}
    \caption{ 
        Performance of character-based text augmentations as an adversarial strategy.
        Each bar represents the proportion of memes that flipped their label after adversarial modifications to the text out of all memes that were classified correctly with no augmentations. 
    }
    \label{fig:text_results}
\end{figure}

We present results on the number of originally correct memes that flipped their label after character-based text augmentations in Figure~\ref{fig:text_results}.
The first thing to observe about these attacks is that they are less effective than image-based attacks.
Attacks that fully disrupt the text (``heavy random augmentations'') without any access to the classifier and attacks with full access to the classifier (``guided beam search'') only cause 25-30\% of originally correctly predicted memes to flip their labels. 
Compare this to the 98\% success rate by full-access image adversaries on grid feature models and the 44-45\% gray box image adversaries on both sets of models.

The different success rates at different levels of access show that adversarial power is also important here.
For example, to control for the risk by an adversary who cannot query the model to an adversary who can query a similar model trained on the same dataset as the target, compare ``light random augmentations'' and ``dataset-access guided beam search.''
As a reminder, the maximum edit distance from the original string is the same in both cases, so human understanding of the message is not impeded.
However, adversaries who get to query even a similar model can craft much more powerful attacks -- they achieve a 16-19\% flip rate, while no-access adversaries only achieve 9-15\%. 

\subsection{Combining Text and Image Adversarial Examples}
\label{sec:exp_both}
\begin{figure}[tb]
    \centering
    \includegraphics[width=\columnwidth]{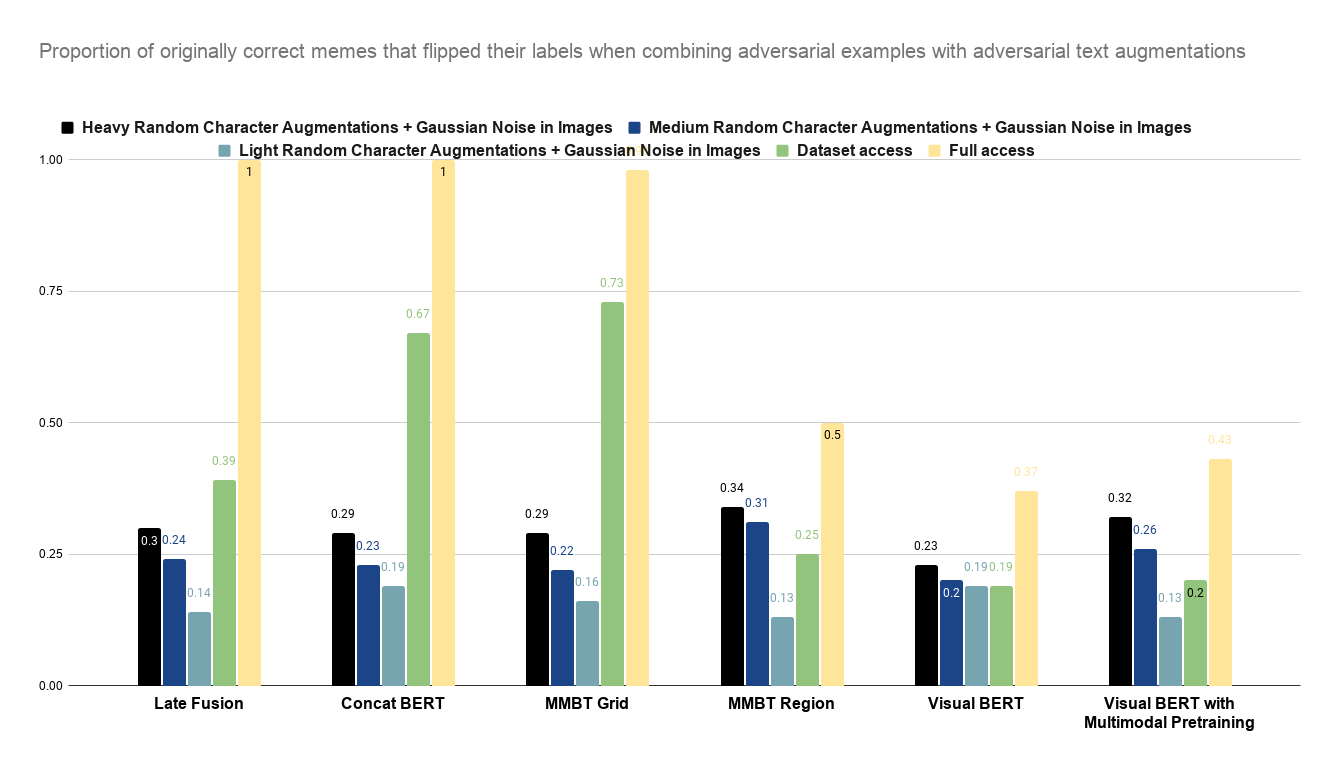}
    \caption{ 
        Average proportion of memes that were originally correct but were flipped by attacks that combine image adversarial examples and adversarial text augmentations.
    }
    \label{fig:both_results}
\end{figure}

We report results in Figure~\ref{fig:both_results} on attacks that affect both modalities. 
All attacks are generated by taking the corresponding adversarial image example and adversarially augmented text by an adversary with the same powers.
Since there is no equivalent for an adversary who possesses the image feature extractor for the text attacks, we omit this category here.
As can be expected, all models perform worse when attacks across modalities are combined. Observe that two models in particular are the worst affected: the concatenation-based ``mid fusion'' ConcatBERT model and the ``early fusion'' grid features-based MMBT model. 
Under gray box assumptions (an adversary with dataset knowledge only), the former is fooled 73\% of the time it used to be correct and the latter is fooled 67\% of the time. 
This may suggest that mid and early fusion models relying on grid features are most vulnerable to attacks that are themselves multimodal, even if they do not stand out in vulnerability under attacks in any one single modality.

\section{Conclusion}
\label{sec:conclusion}
This work shows that multimodal models combining text and image data are vulnerable to attacks even when adversaries do not have access to every piece of their pipeline.
Strong image-based attacks exist regardless of the feature extractor used.
However, it is strictly easier to attack region features-based models as they rely on a publicly available component.
Our work opens exciting new avenues for future research.
For example, to protect against gray box adversaries, defenses should focus on more robust image feature extraction and aim to reduce transferability of adversarial examples from models trained on the same dataset. 

\section*{Acknowledgements}
The authors would like to thank the Facebook AI Red Team, Aaron Jaech, Amanpreet Singh, and Vedanuj Goswami for their help. 
At the University of Washington, Ivan Evtimov is supported in part by the University of Washington Tech Policy Lab, which receives support from: the William and Flora Hewlett Foundation, the John D. and Catherine T. MacArthur Foundation, Microsoft, the Pierre and Pamela Omidyar Fund at the Silicon Valley Community Foundation; he is also supported by the US National Science Foundation (Award 156525).

{\small
\bibliographystyle{ieee_fullname}
\bibliography{egbib}
}

\appendix

\section{Background on Multimodal Classification}
\label{sec:background_multimodal}
There is no one ``best'' way to achieve good performance for multimodal classification. 
Here, we describe the most popular approaches, represented by the baseline models in the Hateful Memes challenge~\cite{kiela2020hateful}.
There is generally a 3 step process:
\begin{enumerate}
    \item \label{mmclassification:step_one} Extract features from the image component
    \item \label{mmclassification:step_two} Process text for classification (In some cases, this includes feature extraction.)
    \item \label{mmclassification:step_three} Make predictions based on image features and processed text
\end{enumerate}
We summarize the choices for each step in Figure~\ref{fig:mm_models} and elaborate in the following two sections.

\begin{table}[tb]
    \centering
    \includegraphics[width=\columnwidth]{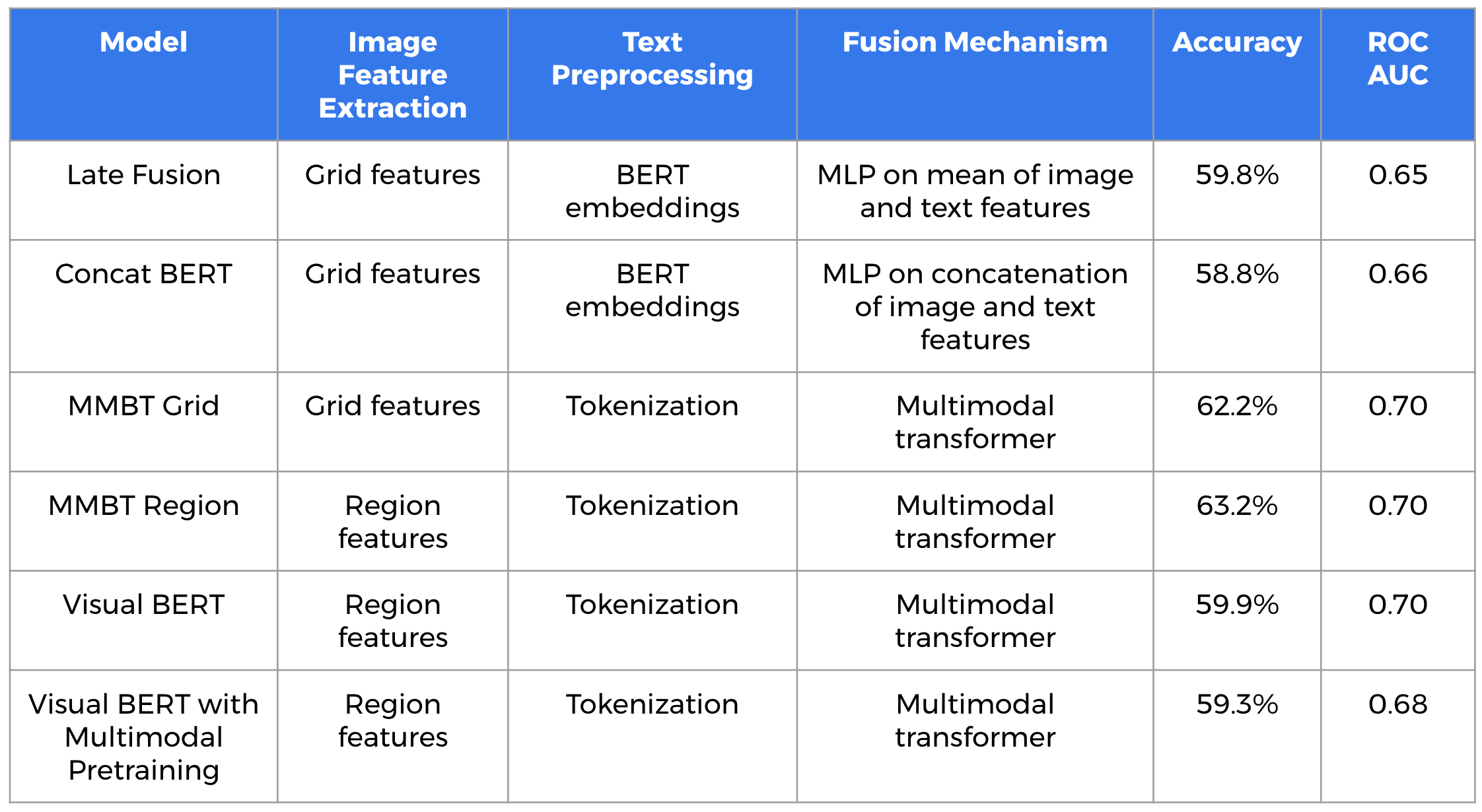}
    \caption{ 
    Overview of the multimodal classification models evaluated and the performance metrics we were able to replicate
    }
    \label{fig:mm_models}
\end{table}

\subsection{Image Feature Extraction}
Step~\ref{mmclassification:step_one} extracts semantic information from images for further processing by other models.
There are two common methods for feature extraction -- grid-based and region-based.
Grid feature extraction uses feature maps from popular convolutional neural networks (CNNs) as image features. 
For example, for a ResNet~\cite{He2016deep} network, the output of the \texttt{res5c} layer is taken as representing the image for downstream multimodal classification.

\begin{figure}[tb]
    \centering
    \includegraphics[width=\columnwidth]{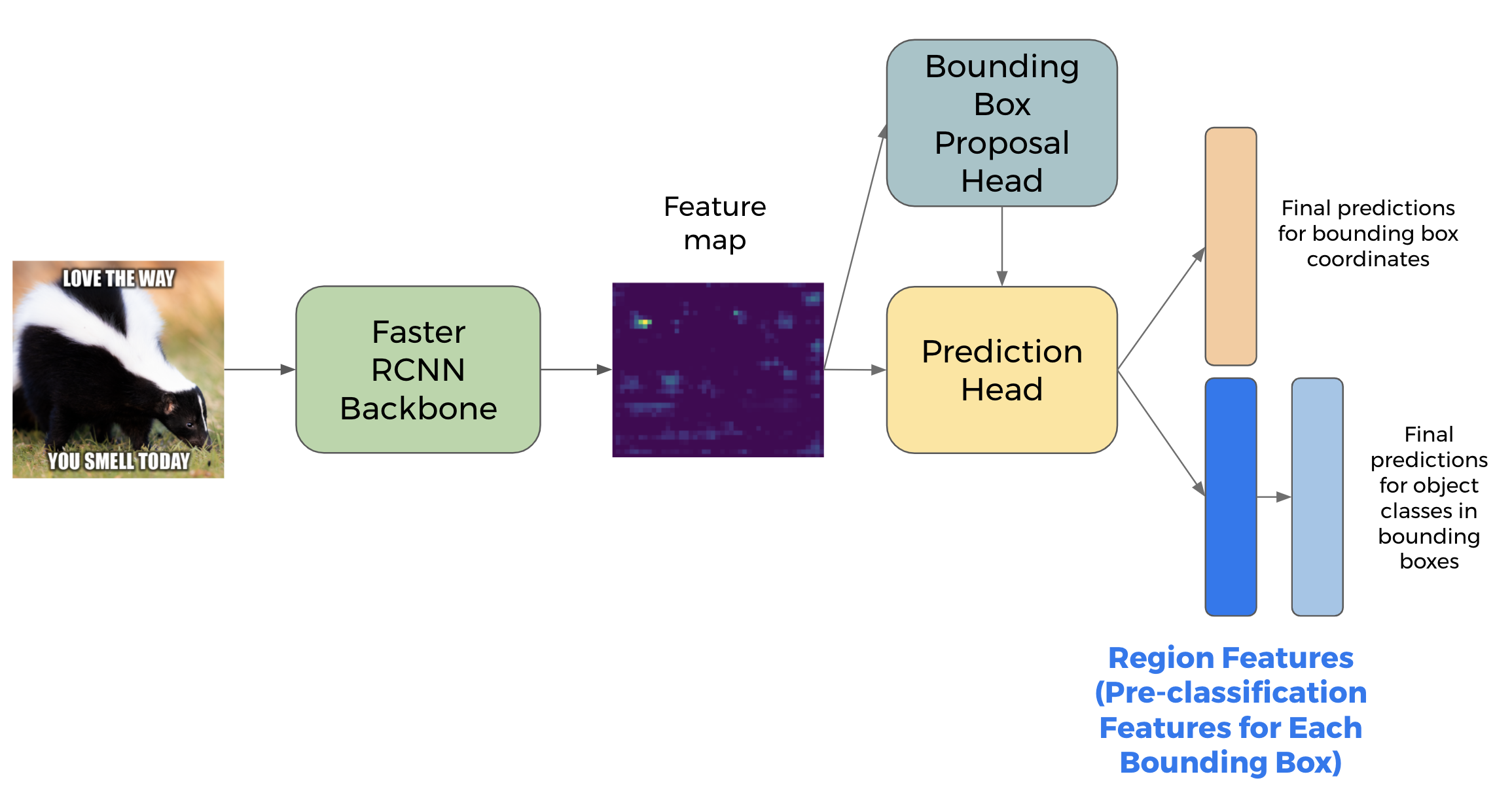}
    \caption{ 
    Overview of region feature extraction from images with Faster R-CNN models~\cite{ren2015faster}.\vspace*{2mm}
    }
    \label{fig:region_features}
\end{figure}

Region feature extraction is more widely used and is critical to our attack design.
The baseline models for the Hateful Memes Challenge use the Faster R-CNN family of models to extract region features. 
Faster R-CNN processes images in three stages which we illustrate in Figure~\ref{fig:region_features}.
First, a backbone network produces feature maps distilling the contents of the input image at several different resolutions.
Then, a branch of the network (the ``proposal head'') uses the feature map to generate rough-estimate proposals for bounding boxes where an object may be located.
These proposals are refined by another branch (the ``prediction head'') to produce exact coordinates and classes for each proposal, also using the feature map.
Finally, non-max suppression is applied so that each object is captured in only one bounding box.
The ``region features'' are the output of the fully connected layer before the softmax classification layer in the prediction head, for each proposed bounding box that survives non-max suppression.

While there is an active debate~\cite{jiang2020defense} in the multimodal research community about which features achieve better performance, there is a significant distinction between the two for adversarial analysis.
Because grid feature models are more lightweight, multimodal models built on top of them are trained end-to-end.
Thus, no off-the-shelf components are used in deployment.
By contrast, region feature models are often too heavy to train end-to-end and publicly available, pretrained models are used without fine-tuning.
This implies different levels of access for the attacker which we describe in more detail in Section~\ref{sec:threat_model}.

The Late Fusion, ConcatBERT and one version of the MMBT model (developed in~\cite{kiela2020hateful} and~\cite{Kiela2019supervised}) use grid features while the Visual BERT model (first introduced in~\cite{Li2019visualbert} and adapted in~\cite{kiela2020hateful})  and another version of MMBT use region features.
All models we work with are implemented in the MMF library~\cite{singh2020mmf} and trained on the Hateful Memes dataset~\cite{kiela2020hateful}.

\subsection{Text Preprocessing and Multimodal Fusion}
The method used for text preprocessing for the text modality is closely tied to the multimodal fusion mechanism used.
It is useful to think of 3 levels of fusion: late, mid, and early.
In late and mid fusion, high-level semantic embeddings are extracted from text in a similar fashion to image feature extraction.
A unimodal text model is used to pre-process this modality.
For example, in the ConcatBERT and Late Fusion models from the Hateful Memes challenge, BERT embeddings are used.
But in other cases, such embeddings could be derived by character-based CNNs and RNNs.
In those situations, the resulting text embeddings are combined with the image features either through concatenation (mid fusion) or averaging (late fusion) and a ``light-weight'' multilayer perceptron (MLP) is used on top.
In early fusion settings, text tokens are instead fed directly into a multimodal transformer in the same way they are fed to unimodal transformers (such as BERT).
To prepare image features for the format required by transformers, affine transformations can be learned (as in MMBT~\cite{Kiela2019supervised}) or engineered to indicate they are image embeddings (as in Visual BERT~\cite{Li2019visualbert}).
In both cases, we treat this preprocessing of the image embeddings as a frozen part of the multimodal transformer that the attacker has no control over -- just like the weights and biases deeper in the transformer.

\section{Threat Model}
\label{sec:threat_model}
We will consider five different adversaries, each distinguished by their knowledge or technical capability of generating attacks.

To begin with, the most powerful adversary we consider is the classic full-access attacker common to the majority of the adversarial examples research literature. 
We assume that they possess the architecture and weights of the exact models used in the deployment of multimodal classification.
They can, therefore, run and backpropagate through those on their own and adapt their attacks as necessary. 
Considering this adversary helps define the bounds of what attacks are possible in the worst case for the system designers.

However, attackers with less knowledge than this may also be of concern and we also seek to understand what they are capable of. 
Consider two important pieces of building multimodal classification: the dataset used to train the models and the image feature extractor used to generate inputs for the image modality.

First, adversaries may possess the dataset used to train the multimodal models even if they do not know which model in particular is being used.
Such data sets are often made public for academic research purposes.
Adversaries seeking to attack a hatefulness classification system can, therefore, certainly train their own multimodal models to guide their attack generation.

Second, attackers are likely to have access to the exact image feature extractor even if they do not have access to either the exact dataset used or the full model. 
As we mentioned in Section~\ref{sec:background_multimodal}, it is a common practice in multimodal classification to use so-called ``region features'' for images.
Those are extracted with publicly available 
\footnote{Implementations and weights of those models are available, for example, at \url{https://github.com/rbgirshick/py-faster-rcnn} and \url{https://github.com/facebookresearch/grid-feats-vqa}.} 
object detectors such as Faster R-CNN~\cite{ren2015faster}.

Finally, the system designer applying multimodal classificaiton may choose to not rely either on public datasets or public models for preprocessing. 
Thus, it is also important to consider adversaries who do not have this level of access.
We differentiate between two possible attackers in this category.
On the one hand, technically savvy adversaries may use public computer vision models to guide their process of generating adversarial examples.
They could, for example, obtain implementations of ImageNet~\cite{deng2009imagenet} classifiers.
On the other hand, adversaries may not have any expertise in machine learning at all. 
In those cases, they can insert noise or augmentations that are not guided by any model at all. 

Thus, the five adversaries we consider are:
\begin{enumerate}
    \item Full-access attackers possess the multimodal model weights and architecture
    \item Dataset-access attackers possess the dataset used to train the multimodal classifiers but not the exact models being used.
    \item Feature extractor-access attackers possess the component of the pipeline used to extract image features.
    \item No-access, technically savvy attackers have machine learning knowledge but no access to any component of multimodal classification.
    \item No-access, low expertise attackers do not have any machine learning knowledge whatsoever.
\end{enumerate}

\section{Further Details on Methodology}
\label{sec:more_methodology}
\subsection{Formulation of the PGD Objective}
The generic adversarial objective for an image $x$, a model $f$, adversarial loss function $L$, and maximum perturbation $\epsilon$ is as follows:

\begin{equation}
    \label{eq:pgd_generic}
    x' = \arg \min_{x} L\left(f(x)\right) \text{ s. t. } |x' - x| \leq \epsilon
\end{equation}

This objective is solved by gradient descent by using the following update rule:
\[
    x_{i+i} = \texttt{Proj}\left(
        x_i - \alpha \nabla_{x_i} L
    \right)
\]
where $\alpha$ is the learning rate and $\texttt{Proj}$ is the projection function on an $\mathcal{L}_\infty$ sphere of radius $\epsilon$ around the original image $x$.
We set $\epsilon=0.1$ and $\alpha=0.05$ for all experiments that we report results on.

We need to further specify three components for equation~\ref{eq:pgd_generic}: the model $f$ used to generate adversarial examples, the form of the loss $L$ (e.g. cross entropy or L2 distance), and the targets used in the adversarial examples generation.
The choice for each of these varies depending on the threat model and what model the adversary has access to.
We summarize our choices for each threat model in Table~\ref{tab:adv_choices}.
We summarize our choices for each level of adversarial power in Table~\ref{tab:adv_choices}.

\begin{table}[tb]
    \centering
    \includegraphics[width=\columnwidth]{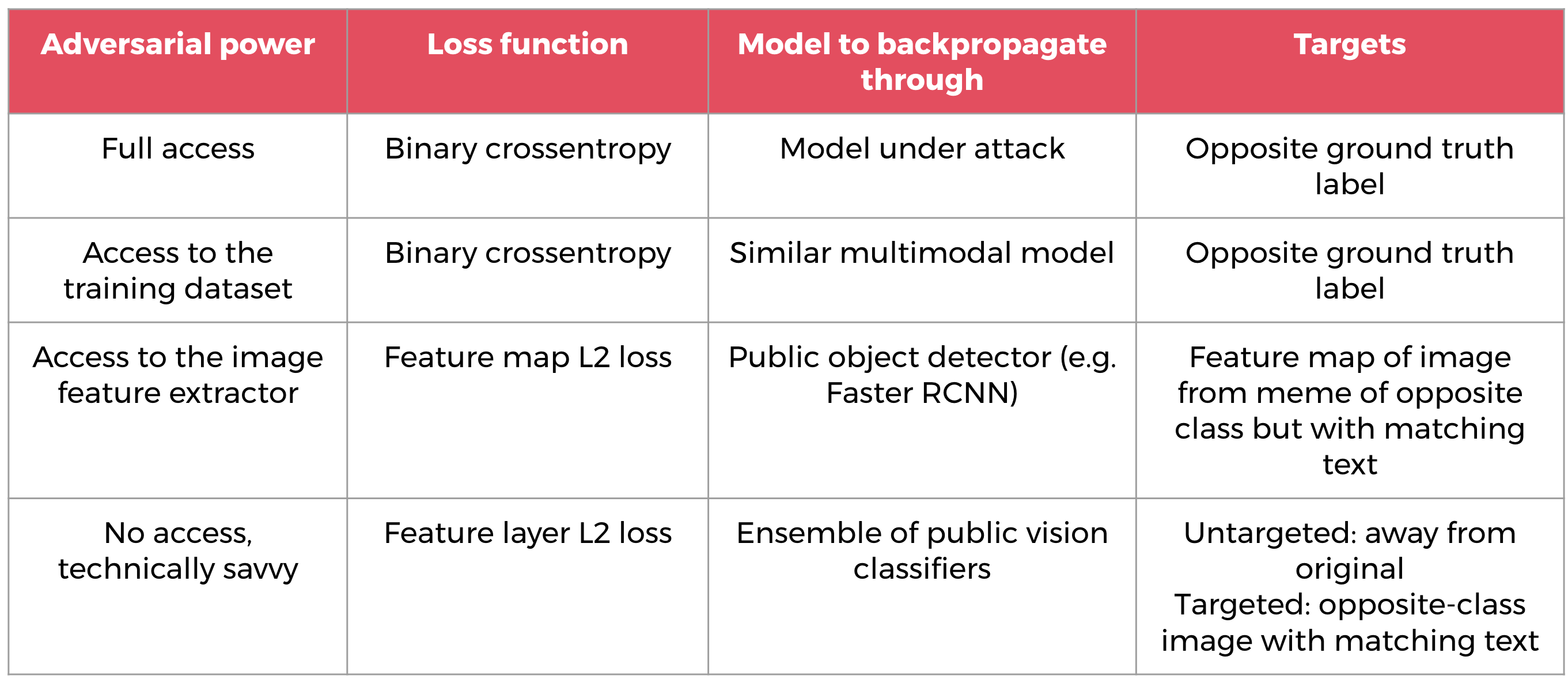}
    \caption{Overview of the attack strategies used to instantiate Eq.~\ref{eq:pgd_generic} for image-based attacks}
    \label{tab:adv_choices}
\end{table}

\subsection{Full-Access and Dataset-Access Attacks}
\label{sec:full_access}
Note that for region feature models, we cannot directly use backpropagation to the input $x$ because object detectors apply non-max suppression to select bounding boxes (and, by extension, the features that become inputs to the multimodal fusion mechanism).
Moreover, gradient descent is likely to be unstable as slight perturbations in each step cause disproportionate changes in the bounding box proposals and associated features.
Instead, we employ the following two-step procedure for Faster R-CNN features:
\begin{enumerate}
    \item \label{list:region_whitebox_one} Generate a single adversarial vector $y'$ for the multimodal transformer. In this case, $f$ is the multimodal transformer with the text input fixed and $L$ is the binary cross-entropy function with adversarial labels. $y'$ is prepended to the sequence of image feature input vectors while all other vectors are held constant.
    \item \label{list:region_whitebox_two} Generate an adversarial example $x'$ such that the Faster R-CNN detector produces that vector $y'$ for all proposed bounding boxes. In this case, $f$ is the Faster R-CNN classification head with the classification layer removed and $L = \sum_{i}||f(x')_i - y'||_2$ for each feature vector $f(\cdot)_i$ output by the Faster R-CNN classification head. 
\end{enumerate}

We note that stronger adversaries in the full-access scenario are likely possible. 
Our exploration only focused on adversaries that first generate a single adversarial feature to be included in those produced by the region feature extractor and then produce adversarial images that output that feature.
This is because we cannot backpropagate end-to-end with region feature extractors (they include a non-differentiable non-max suppression step). 
While stronger adversaries with full access were outside the scope of this work, those are likely to exist, so the 24\% number should be treated as a lower bound.

\subsection{Feature Extractor-Access Attacks}
\label{sec:matching_text}
To use their access to an off-the-shelf feature extractor (used in multimodal classification) to carry out an attack, adversaries can instantiate Eq.~\ref{eq:pgd_generic} with two pieces: choosing what features the adversarial images should generate and designing an appropriate loss function.

Recall that the Hateful Memes Challenge Dataset contains non-hateful confounders that were created by hand by replacing images of hateful memes with such that make the overall message of the meme non-hateful even while leaving the text unchanged. 
(See Figure~\ref{fig:hateful_memes_examples}.)

To produce those features, we propose an adversarial loss function that targets the feature map in the detector pipeline that is used to produce both proposal boxes and features. 
Recall from Appendix~\ref{sec:background_multimodal} that the region features produced for a given image are computed from the feature map layer of a Faster R-CNN object detector.
Thus, if an adversarially modified hateful image produces a feature map corresponding to a non-hateful image, the region features computed will correspond to the non-hateful image. 
Therefore, we design our loss function so that it penalizes the distance between the feature map of a desired non-hateful target image and the adversarial one we are optimizing.

Formally, let $x$ be the image of a hateful meme and let $y$ be the image of a meme with the same text but an image that makes it benign. 
Further let $f$ be the Faster R-CNN feature map layer (e.g., c4).
Then, in Eq.~\ref{eq:pgd_generic}, $L = ||f(x) - f(y)||_2$. 
The same loss function holds if $x$ is non-hateful and $y$ is its hateful counterpart with matching text. 

In the Hateful Memes test set, we found 483 memes that had a corresponding coutnerpart with matching text but the opposite ground truth label.
Experimental results with models with region features and feature-extractor access, therefore, report the success rate as a proportion of 483.

\subsection{No-Access Attacks by Technically Savvy Adversaries}
Adversarial examples are generated with a white-box attack that averages the gradient from $n$ models. 
For models $f_1, ..., f_n$, we modify the PGD objective as follows:
\begin{equation}
    \label{eq:pgd_ensemble}
    x' = \arg \min_{x} \frac{1}{n} \sum_{i=1}^n L\left(f_i(x)\right) \text{ s. t. } |x' - x| \leq \epsilon
\end{equation}
In all cases, $f_i$ is taken to mean the output of the final convolutonal feature map in the corresponding network (e.g. \texttt{res5c} in ResNet). 
We work with ResNet-152~\cite{He2016deep}, ResNext-50~\cite{xie2017aggregated}, Inception-v3~\cite{szegedy2016rethinking}, VGG-16~\cite{simonyan2014very}, and DenseNet~\cite{huang2017densely}. 
We further introduce two versions of Equation~\ref{eq:pgd_ensemble}: 
\begin{itemize}
    \item In \textit{untargeted} attacks for original image $x$ and adversarial images $x'$, we set $L = - ||f(x) - f(x')||_2$ to create adversarial images that shift the feature map as far away from its original as possible.
    \item In \textit{targeted} attacks, we set $L = ||z - f(x')||_2$ for some target feature map $z$. Feature maps are selected as in Section~\ref{sec:matching_text}.
\end{itemize}

\section{Details of Algorithm for Text-Based Adversarial Augmentations}
\label{app:adversarial_text}
\subsection{Guided Adversarial Text Augmentations}
\label{sec:guided_adv_text}
Input to text models such as BERT is discrete, so using gradient-based approaches is not possible.
However, we can still use queries to the model to guide a search for adversarial text augmentations.
Therefore, we adapt beam search techniques (such as those used in~\cite{ebrahimi2017hotflip}), to the multimodal scenario.

Our adversarial search algorithm works as follows.
For a given string we want to adversarially augment, we apply a set of character-based augmentations.
At each step of the beam search, we generate multiple different variants at random and only a minimal character augmentation is applied to each one (such as only replacing one letter).
We discard augmented strings whose edit distance from the original string exceeds a threshold $\tau$.
What remains is the ``candidate'' set.
We then rank each member of the candidate set according to how big of a drop in confidence on the correct class for the meme under attack it causes.
If the drop by any candidate is big enough to cause an error in classification, we stop there and return the successful augmented string.
If none of the candidates causes a classification mistake, we select the top $k$ to remain in the beam.
Then, we repeat this process by generating multiple randomly augmented candidates for every string in the beam, selecting those under the edit distance threshold, and picking the top $k$ for the next iteration of the beam search.

In white-box scenarios, we use the model under attack to rank the candidates.
In light gray-box scenarios, we use any other multimodal model to perform the ranking.

For heavy augmentations, we cap the edit distance at $\tau=0.5$; for medium augmentations, we cap the edit distance at $\tau=0.2$; and for light ones, we set $\tau=0.07$.

\subsection{Random Adversarial Text Augmentations}
\label{sec:unguided_adv_text}
Just as in the image domain, it is useful to generate adversarial text that does not require queries or other access to the model under attack.
We use the same set of character-based augmentations as in the guided scenario, but perform random search to pick them instead of a guided beam search.
We study 3 levels of adversarial modifications:
\begin{itemize}
    \item Light: These are selected so that the maximum edit distance from the original matches that of the adversarial strings generated by the beam search.
    \item Medium: These are selected so that the average edit distance from the original string matches that of the beam search.
    \item Heavy: We do not restrict the edit distance and allow for maximum corruption of the text string. The text in this case carries no human-interpretable message and violates our attack requirement but this is a useful ``upper bound'' on the effectiveness of text-based attacks on multimodal classification.
\end{itemize}

\end{document}